\documentclass[letterpaper]{article} 
\usepackage{aaai25}  
\usepackage{times}  
\usepackage{helvet}  
\usepackage{courier}  
\usepackage[hyphens]{url}  
\usepackage{graphicx} 
\usepackage{natbib}  
\usepackage{caption} 
\usepackage{algorithm}
\usepackage{algorithmic}

\usepackage{newfloat}
\usepackage{listings}
\DeclareCaptionStyle{ruled}{labelfont=normalfont,labelsep=colon,strut=off} 
\floatstyle{ruled}
\newfloat{listing}{tb}{lst}{}
\floatname{listing}{Listing}
\usepackage{amssymb}
\usepackage{booktabs}
\usepackage{amsmath}

\title{CylinderPlane: Nested Cylinder Representation for 3D-aware Image Generation}
\author{
    Ru Jia~~~
    Xiaozhuang Ma~~~
    Jianji Wang\textsuperscript{\textasteriskcentered}~~~
    Nanning Zheng
}
\affiliations{
    Institute of Artificial Intelligence and Robotics, Xi’an Jiaotong University\\
}

\usepackage{bibentry}

\begin{document}


\twocolumn[{%
	\renewcommand\twocolumn[1][]{#1}%
	\maketitle
	\vspace{-18mm}
	\begin{center}
		\centering
		\captionsetup{type=figure}
		\includegraphics[width=\textwidth]{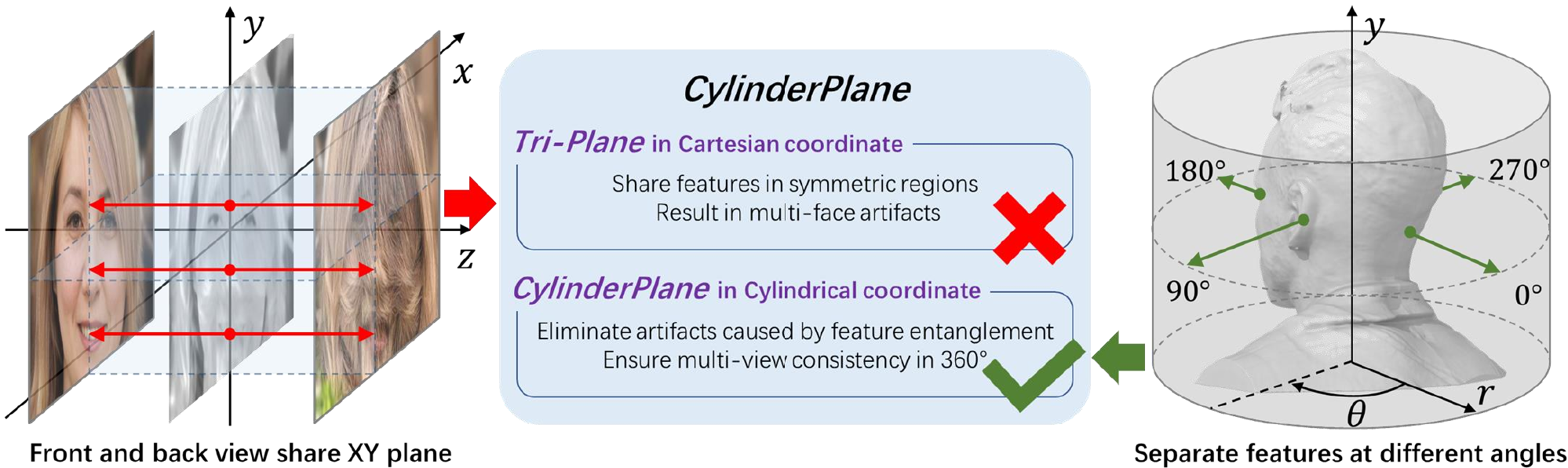}
		\captionof{figure}{\textbf{Overview of CylinderPlane Representation.} The left section illustrates the limitations of the traditional Tri-plane representation, where feature entanglement occurs in symmetrical regions, leading to multi-face artifacts. The right section demonstrates the proposed CylinderPlane representation, which leverages the Cylindrical Coordinate System to separate features at different angles, effectively eliminating the multi-face artifacts and ensuring consistent 360$^\circ$ image synthesis.}
		\label{fig:teaser}
	\end{center}%
}]

\begin{abstract}
	While the proposal of the Tri-plane~\cite{chan2022efficient} representation has advanced the development of the 3D-aware image generative models, problems rooted in its inherent structure, 
	such as multi-face artifacts caused by sharing the same features in symmetric regions, limit its ability to generate 360$^\circ$ view images.
	In this paper, we propose \textit{CylinderPlane}, a novel implicit representation based on \textit{Cylindrical Coordinate System}, to eliminate the feature ambiguity issue and ensure multi-view consistency in 360$^\circ$.
	Different from the inevitable feature entanglement in Cartesian coordinate-based Tri-plane representation, the cylindrical coordinate system explicitly separates features at different angles, allowing our cylindrical representation possible to achieve high-quality, artifacts-free 360$^\circ$ image synthesis.
	We further introduce the nested cylinder representation that composites multiple cylinders at different scales, thereby enabling the model more adaptable to complex geometry and varying resolutions.
	The combination of cylinders with different resolutions can effectively capture more critical locations and multi-scale features, greatly facilitates fine detail learning and robustness to different resolutions.
	Moreover, our representation is agnostic to implicit rendering methods and can be easily integrated into any neural rendering pipeline.
	Extensive experiments on both synthetic dataset and unstructured in-the-wild images demonstrate that our proposed representation achieves superior performance over previous methods.
	
\end{abstract}
\section{Introduction}

\indent
3D generative models, which aim at generating 3D representation from either single/multiple image input or gaussian noises, has long been of great concern within the realms of computer vision and graphics.
These models hold immense potential for a variety of applications, including the game production, telepresence, and virtual/mixed reality~\cite{lu2023tf, lu2024mace,gu2021stylenerf}. 

The rise of Neural Radiance Fields (NeRFs) has spurred numerous approaches for generating 3D scenes through implicit radiance field representations~\cite{mildenhall2021nerf,deng2022gram}, achieving impressive photorealistic rendering quality.
Nevertheless, the typical NeRF framework employs large multi-layer perceptrons (MLPs) to parameterize the radiance field, requiring a substantial number of forward passes during volumetric rendering. This computation-intensive process becomes a bottleneck in applications where speed is critical, such as real-time rendering or GAN-based training scenarios.
To address this inefficiency, a range of acceleration techniques have been introduced~\cite{fridovich2022plenoxels,muller2022instant,chen2022tensorf}.
Among these solutions, the Tri-plane representation~\cite{chan2022efficient} stands out for its balance between speed and detail. By projecting 3D points onto three orthogonal planes, it enables more efficient radiance queries while still capturing fine-grained geometric and texture details.

However, the Tri-plane representation also presents inherent limitations rooted in its planar decomposition and orthogonal projection scheme.
Firstly, the issue arises from feature overlap in symmetrical regions, since orthogonal projections onto predefined Cartesian planes naturally cause different sides of an object to share the same feature samples. This leads to multi-face artifacts commonly referred to as the Janus problem.
As shown in the left part of Figure~\ref{fig:teaser}, both front and back views are generated from identical features on the XY-plane, resulting in erroneous projections where parts of the front appear on the back, breaking the correct view-dependent rendering.
These artifacts severely affect the 3D consistency across wide viewing angles, reducing the sense of realism and immersion in interactive 3D scenes.
Secondly, the use of only three axis-aligned planes (XY, XZ, YZ) limits the system’s ability to capture fine-grained geometry, especially for diagonal structures or curved surfaces that do not align well with the planes.
This constraint can lead to geometric distortions or detail omission.
Additionally, the Tri-plane approach suffers from resolution dependency: the expressiveness of the generated 3D scene is restricted by the feature plane resolution, causing degradation when adapting to scenes at varying levels of detail.

To overcome these challenges, we introduce \textit{CylinderPlane}, a new implicit representation grounded in the \textit{Cylindrical Coordinate System}, aimed at generating high-quality, 3D-consistent outputs across multiple views.
Our method addresses two major limitations of the traditional Tri-plane framework.
First, unlike Cartesian-based projections that inherently duplicate features in symmetric regions, causing ambiguity and the well-known multi-face artifact~\cite{yu2022pvserf}, the cylindrical coordinate system naturally separates angular information.
As illustrated in the right part of Figure~\ref{fig:teaser}, this design explicitly disentangles features along different azimuthal directions, effectively eliminating multi-face artifacts and ensuring consistent synthesis over full 360$^\circ$ viewpoints.
Second, to further enhance geometric expressiveness and resolution adaptability, we propose a nested cylinder mechanism.
By stacking multiple cylindrical feature planes at different radii, we enable the model to capture features at multiple spatial scales.
Unlike the Tri-plane approach, which relies on three fixed-resolution orthogonal planes, our nested cylinders provide continuous angular coverage and concentrate sampling in more informative regions.
This structure significantly improves the model’s ability to reconstruct intricate shapes and detailed textures.
Additionally, the variable radii introduce natural multi-scale capacity, allowing the representation to flexibly handle scenes of diverse resolutions and complexities.

In summary, the contributions of this paper is three-folds:
\begin{itemize}
	\item We propose a novel implicit representation based on Cylindrical Coordinate System. This representation addresses the limitations of traditional Tri-plane representation and can be seamlessly integrated into various neural rendering pipelines.
	\item We propose a multi-scale nested cylinder planes approach, which improve the model's capability to model complex geometric details and adapt to various scene resolutions.
	\item To facilitate the 3D full head generation, we build a panoramic head images dataset using an automatic pipeline. This dataset will be released to the community at a later stage.
\end{itemize}

\section{Method}

\begin{figure*}[!t] \centering
    \includegraphics[width=\linewidth]{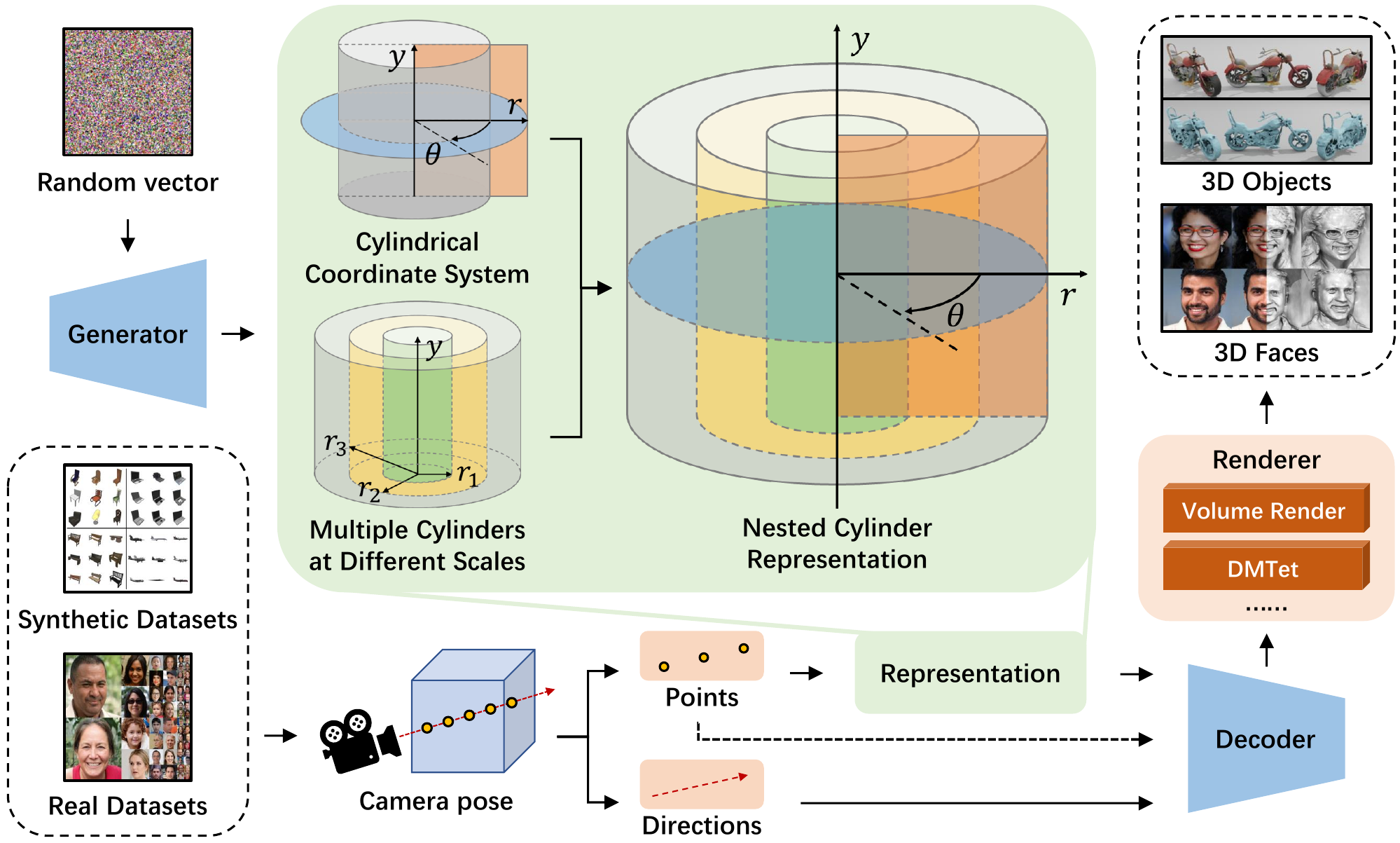}
    \caption{\textbf{Overview of the CylinderPlane pipeline.} A random vector is fed into the StyleGAN-like generator, outputs several planar feature maps which are projected into the \textit{Cylindrical Coordinates System}. The projected cylinder planes are organized as nested cylinders at different scales, which is akin to a ``Swiss Roll". This Nested Cylinder Representation is versatile and can be integrated into various neural renderers, allowing for the creation of 3D-aware outputs, such as 3D faces or objects.} 
    \label{fig:pipeline}
    \vspace{-4mm}
\end{figure*}
\begin{figure}[!t] \centering
    \includegraphics[width=\linewidth]{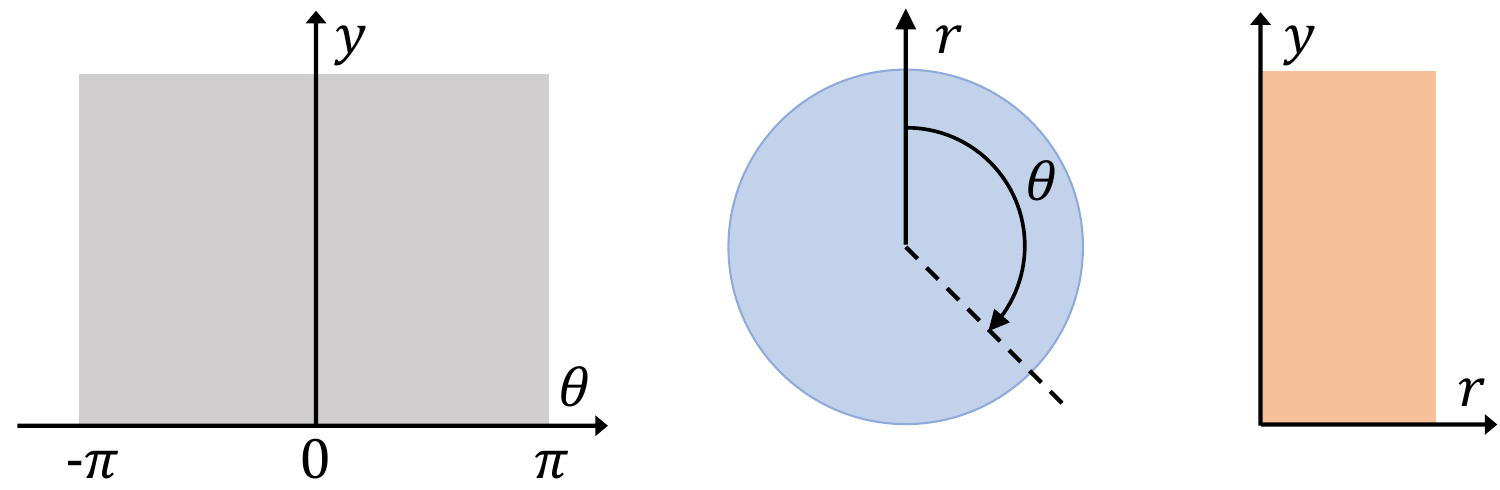}
    \caption{\textbf{Illustration of the three cylindrical planes.}} 
    \vspace{-5mm}
    \label{fig:cylinder}
\end{figure}

We propose the \textit{CylinderPlane} representation, introducing a new paradigm for 3D generative modeling.
As shown in Figure~\ref{fig:pipeline}, starting from a Gaussian noise vector, a generator, such as a StyleGAN-like or diffusion-based model, produces a set of 2D feature planes.
These planes are then mapped into the cylindrical coordinate system, effectively “rolling up” the planar features to form \textit{cylinder planes}, as visualized in Figure~\ref{fig:cylinder}.
To further enhance spatial coverage and multi-scale detail, we arrange several cylinder planes with varying radii and orientations, constructing a Nested Cylinder structure that resembles a ``Swiss Roll''.
This representation is highly flexible and can be directly integrated into different neural rendering frameworks.
During the rendering process, 3D points within the camera frustum dynamically sample features from the nested cylinder planes, which are subsequently decoded into rendering attributes, such as color and density in volumetric rendering pipelines. More technical details are provided in the following sections.

\indent
In the following sections, we begin by presenting the preliminaries and the motivation behind our proposal for a novel implicit representation.
Then, in Section 2, we provide a detailed formulation of the cylinder plane representation based on the Cylindrical Coordinate System.
At last, Section 3 extends this representation into multi-scale nested cylinder planes and elucidates the underlying principles.

\subsection{Preliminary and Motivation}
\label{sec:motivation}
Given a set of unstructured in-the-wild images with estimated camera poses or a set of rendered images from synthetic objects meshes with ground-truth camera poses, the 3D generative models learn a distribution over 3D objects or scenes behind the in-the-wild images, allowing for the generation of novel 3D structures and their subsequent rendering into 2D images.

\indent
Specifically, given a paired image and its corresponding camera pose $(\mathcal{I}, \mathcal{K})$, a 3D generative model that outputs a 3D representation \( \mathbf{X} \) from a latent code \( \mathbf{z} \):
\[
\mathbf{X} = G(\mathbf{z})
\]
where \( G \) is the generator network, and \( \mathbf{z} \) is sampled from a prior distribution, typically \( \mathbf{z} \sim \mathcal{N}(0, \mathbf{I}) \), a standard normal distribution.
The 3D points $x \in \mathbb{R}^3$ within the camera frustum defined by the camera pose $\mathcal{K}$ then acquire their features from this 3D representation \( \mathbf{X} \). These acquired features are finally decoded into radiance properties and rendered into RGB images using differentiable rendering techniques.

\indent
The 3D representation \( \mathbf{X} \) can take various forms, such as a vanilla MLPs~\cite{mildenhall2021nerf}, voxel feature grid~\cite{fridovich2022plenoxels}, or Tri-planes. 
For instance, in the case of Tri-plane representation, \( G \) outputs three axis-aligned orthogonal feature planes, each with a resolution of $N \times N \times C$, with $N$ being spatial resolution and $C$ the number of channels. Then a 3D position $x \in \mathbb{R}^3$ acquires its feature by projecting itself onto each of the three feature planes, retrieving the corresponding feature vector ($F_{xy}$, $F_{xz}$, $F_{yz}$) via bilinear interpolation, and aggregating the three feature vectors through summation.

\indent
Despite the efficiency of the Tri-plane representation, it has two major drawbacks: 
First, its planar characteristics and orthogonal projection cause feature entanglement in symmetrical areas, leading to multi-face artifacts, known as the Janus problem. 
Second, the reliance on three fixed-resolution orthogonal planes (XY, XZ, YZ) limits the ability to capture of complex geometric details and reduces robustness across resolutions. 
These limitations motivates us to design a novel representation that addresses these issues, and the following section will elaborate our design choices.

\begin{figure*}[!t] \centering
    \includegraphics[width=\linewidth]{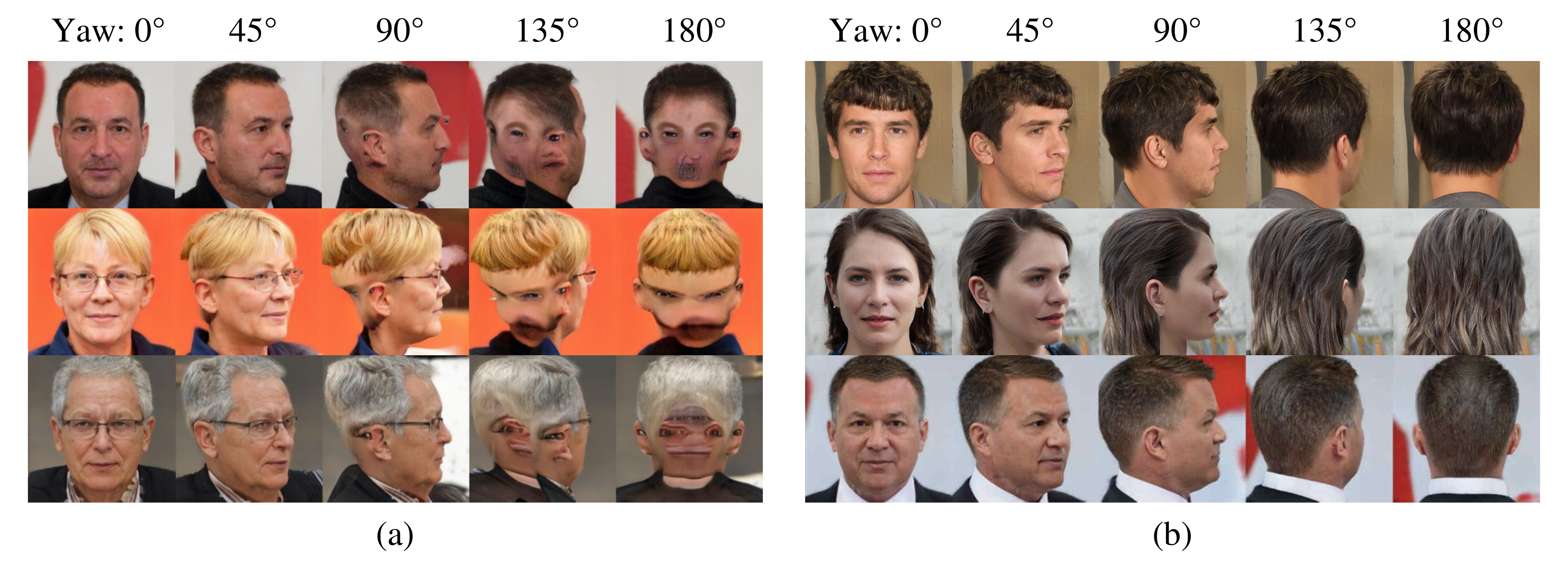}
    \vspace{-7mm}
    \caption{\textbf{Visual comparison with PanoHead~\cite{an2023panohead}}. (a) PanoHead, (b) Ours. The results of PanoHead suffer from the obvious multi-face artifacts, whereas our results exhibit strong 3D consistency.} 
    \label{fig:head_comp}
    \vspace{-5mm}
\end{figure*}

\subsection{Cylinder Plane based on Cylindrical Coordinate System}
\label{sec:cylindrical}
The Tri-plane representation offers an efficient and compact method for 3D-aware generation. 
However, in practical scenarios, methods based on Tri-planes frequently encounter multi-face artifacts, especially when synthesizing wide field-of-view images.
This problem is exacerbated when the training images have an imbalanced camera distribution, causing dominant cameras to disproportionately affect the Tri-plane features.
While one aspect of this problem can be attributed to the biased supervision from imbalanced training data, a more significant factor is the inherent Cartesian projection of the Tri-plane representation, which inevitably leads to feature entanglement at symmetric positions. For instance, as illustrated in Figure~\ref{fig:teaser}, the front and back views share features at the identical locations on the XY plane.

\indent
To address this issue, we propose a novel cylindrical coordinate representation that explicitly separates features from different angles, thereby eliminating undesirable artifacts. 
Specifically, as illustrated in Figure~\ref{fig:pipeline}, we redefine the position of a point within the volume based on a Cylindrical Coordinate System as $(\theta, r, y)$, and re-express the feature planes as a cylindrical plane ${\theta}y$, a circular plane $r\theta$, and a rectangular plane $yr$, as depicted in Figure~\ref{fig:cylinder}. Similar to the Tri-plane approach, the neural radiance density and color of a point within the volume can be obtained by projecting its cylindrical coordinates onto the three feature planes $F_{{\theta}y}, F_{r\theta}, F_{yr}$, and summing the bilinearly interpolated features from these planes.
In practice, since unfolding the $F_{r\theta}$ and $F_{yr}$ planes along $r$-axis would only occupy half of the feature map from the StyleGAN-like generator, we re-parameterize them as two square planes to fully exploit the generative ability.

\paragraph{CylinderPlane Boundary Regularization}
Due to the numerical discontinuity at $\theta = -\pi$ and $\theta = \pi$ on the cylindrical plane ${\theta}y$, the resulting output suffers from mismatch artifacts or high-frequency noise in the seam regions. 
To address this issue, we introduce two regularization terms at $\theta = -\pi$ and $\theta = \pi$, guiding the two sides of the seam region to converge. First, we apply a constraint by calculating the difference between features at $\theta = -\pi$ and $\theta = \pi$. Second, we apply feature smoothing at the seam region to further alleviate the high-frequency noises. (Details can be found in the Supplementary Material) By combining constraints with smoothing, we effectively eliminate the artifacts in the seam regions.

\subsection{Multi-scale Nested Cylinder Planes}
\label{sec:nested}
The Tri-plane method's reconstruction relies entirely on three orthogonal feature planes, making their structure and resolution crucial for capturing geometric details. However, because these planes are fixed in orthogonal directions and MLPs typically learn low-frequency structures, high-frequency details in complex scenes might be missed or poorly represented. This issue is particularly noticeable in scenes with complex surfaces or significant depth differences, like curved edges or oblique surfaces. Additionally, the fixed resolution of these planes limits their ability to capture fine details at high resolutions and may introduce noise in low-resolution scenes.

\indent
To address these limitations, we further employ a combination of nested cylinders at different scales, which allows sampling from all directions and more critical positions, significantly enhancing the learning of intricate details. 
Specifically, in addition to the three planes of the Cylindrical Coordinate System ($F_{{\theta}y}, F_{r\theta}, F_{yr}$), we introduce $N \in \mathbb{Z}$ cylindrical surfaces with different radii nested within each other:
\begin{equation}
\begin{aligned}
	F_{{\theta}y} = \{ F^{r_0}_{{\theta}y}, F^{r_1}_{{\theta}y}, ..., F^{r_N}_{{\theta}y} \}, \\
	r_0 < r_1 < ... < r_N.
\end{aligned}	
\end{equation}
Due to the varying resolutions of the cylindrical surfaces with different radii, the multi-layer cylindrical combination can capture multi-scale features of the scene, thus achieving robustness across scenes with different resolutions.

\subsection{Integration with Neural Rendering Pipelines}
As shown in Figure~\ref{fig:pipeline}, the Nested Cylinder Representation can be integrated into various rendering pipelines, as long as they are differentiable. Here, we present two exemplar renderers: the volume renderer and the DMTet~\cite{shen2021dmtet}-based differentiable mesh rasterizer. In the volume renderer, for any points on the camera rays, our CylinderPlane outputs its feature representation which will be decoded into color and density for volume rendering. In the DMTet-based mesh rasterizer, the CylinderPlane is incorporated to model the texture field that produce colors for mesh surface points.

\section{Experiments}
Experiments are performed to validate the effectiveness of the proposed CylinderPlane representation. Firstly, we evaluate its performance on the 3D full-head synthesis task, comparing the synthesized results with those from existing methods.

\subsection{Experiments on In-the-wild Head Images}

\paragraph{Datasets}
A key obstacle in generating high-fidelity 3D full-head models covering $360^\circ$ views is the scarcity of publicly available, high-quality panoramic head datasets.
To mitigate this limitation, we construct a comprehensive Full-Head dataset by integrating and processing images from FFHQ~\cite{karras2019style}, LPFF~\cite{wu2023lpff}, and K-hair~\cite{kim2021k}. This dataset serves as the foundation for both training and evaluating our synthesis framework.
We develop an automated data processing pipeline that filters out low-quality samples, removes images containing multiple faces, and restores occluded regions in K-hair using mosaic-based completion.
Through this pipeline, we obtain nearly 300K high-resolution full-head images spanning complete $360^\circ$ views.

\paragraph{Baseline and Implementation Detail}
We choose the state-of-the-art 3D full-head synthesis method, PanoHead, as the comparing baseline method. 
Following PanoHead, we evaluate all the FID-front, FID-back and FID-all~\cite{heusel2017gans} metrics to numerically compare the results.

\begin{table}[!t]
	\caption{\textbf{Numerical comparison of 3D full-head synthesis methods.}}
	\label{tab:comp_head}
	\centering

	\begin{tabular}{ccc}
	\toprule
	& PanoHead~\cite{an2023panohead}  & Ours  \\ 
	\midrule
	FID-front $\downarrow$    & 5.94  & \textbf{5.22}    \\
	FID-back $\downarrow$   & 51.61  & \textbf{40.83}   \\
	FID-all $\downarrow$   & 5.98  & \textbf{5.15}   \\
	\bottomrule
	\end{tabular}
\vspace{-4mm}
\end{table}

\paragraph{Results}
The numerical and visual results are presented in Table~\ref{tab:comp_head} and Figure~\ref{fig:head_comp}, respectively. It can be observed that PanoHead exhibits vbvious Janus artifacts, while our method demonstrates strong 3D consistency, particularly in the back of the head regions. The numerical results in Table~\ref{tab:comp_head} further reflect this, as our FID-back score is evidently better than that of PanoHead, proving the effectiveness of our design.

\section{Conclusion}
We introduce \textit{CylinderPlane}, an implicit 3D representation constructed within the \textit{Cylindrical Coordinate System}, specifically designed for high-fidelity, multi-view consistent generative modeling.
Unlike conventional Tri-plane methods, our approach mitigates the problem of feature entanglement in symmetric regions by leveraging a nested cylindrical structure.
This multi-scale design enables better modeling of intricate geometries while maintaining flexibility across different resolution levels.

Comprehensive experiments on both synthetic and real-world datasets demonstrate the superiority of our method compared to existing baselines.
Looking ahead, potential extensions include adapting CylinderPlane for dynamic scene generation and further improving its efficiency in high-resolution or real-time rendering pipelines.

%

\bibliography{aaai25}

\end{document}